\documentclass[conference]{IEEEtran}
\IEEEoverridecommandlockouts
\usepackage{cite}
\usepackage{amsmath,amssymb,amsfonts}
\usepackage{algorithmic}
\usepackage{graphicx}
\usepackage{textcomp}
\usepackage{xcolor}
\def\BibTeX{{\rm B\kern-.05em{\sc i\kern-.025em b}\kern-.08em
    T\kern-.1667em\lower.7ex\hbox{E}\kern-.125emX}}
\begin{document}

\title{Improving Cancer Imaging Diagnosis with Bayesian Networks and Deep Learning: A Bayesian Deep Learning Approach\\
}

\author{\IEEEauthorblockN{Pei Xi (Alex) Lin}
\IEEEauthorblockA{\textit{Faculty of Mathematics} \\
\textit{University of Waterloo}\\
Waterloo, Canada \\
pxlin@uwaterloo.ca}
}

\maketitle

\begin{abstract}
With recent advancements in the development of artificial intelligence applications using theories and algorithms in machine learning, many accurate models can be created to train and predict on given datasets. With the realization of the importance of imaging interpretation in cancer diagnosis, this article aims to investigate the theory behind Deep Learning and Bayesian Network prediction models. Based on the advantages and drawbacks of each model, different approaches will be used to construct a Bayesian Deep Learning Model, combining the strengths while minimizing the weaknesses. Finally, the applications and accuracy of the resulting Bayesian Deep Learning approach in the health industry in classifying images will be analyzed. 
\end{abstract}

\section{Introduction}
Cancer being one of the most intimidating illnesses coupled with its unpredictable nature and impact on lives has made it one of the top priorities for research. It is the leading cause of premature death in 57 countries. Although 80\% of countries had plans for cancer research, the number of correctly financed and resource-allocated facilities in these countries is still at the bare minimum \cite{b1}. Hence, it is crucial to develop the role of AI in cancer treatment. This project aims to explore and analyze the potential of the combination of artificial intelligence models (deep learning and Bayesian network) to identify the risks of developing cancer in individual patients.

\section{Background}

Computer scientists have been attempting to transition the globe into an artificially intelligent world during the era of technology. Increased research effort has been put into the development of artificial intelligence (AI). Related research on arXiv has reported the number of Artificial Intelligence articles has increased roughly from 6000 to 35,000, almost 7 times \cite{b2}. In the foreseeable future, AI is expected to be used in many industrial fields to enhance work quality and efficiency. Medical researchers have also noticed the trend and capabilities of AI and have begun to adopt it in medical research. However, even though studies in different medical fields have shown that AI technology is successful and effective, only a small portion of AI tools are being used in current medical procedures \cite{b3}. A new approach will be discussed below as another idea to use AI contributions in medical research, specifically, cancer research.  

\subsection{Deep Learning}

\begin{figure}[htbp]
\centerline{\includegraphics{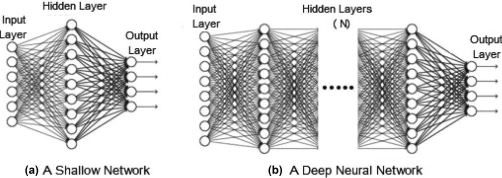}}
\caption{Structure of Deep Learning Model.}
\label{dnn}
\end{figure}

Deep learning (DL) is a type of predictive model in machine learning applications. DL is a machine-learning approach to making predictions through processing data in multiple layers of neural networks. As shown in Fig.~\ref{dnn}, the name of DL comes from its structure of multiple layers and stages used for processing input data and making predictions \cite{b4}. Using a Restricted Boltzmann machine (RBM), the input layer consists of the inputs and a bias, the purpose of this layer is to let the model learn all the input features. Then each subsequent layer is calculated by applying weights and activation function evaluations to the output of previous layers. \cite{b5}. A sample construction of a DL model consists of a learning rate, a forward propagation process to compute the error of predictions, a backward propagation to optimize the problem by computing gradients of the functions, and a stochastic gradient descent function to help reach an optimal minimum. \cite{b6} DL's multi-layered characteristic allows it to process a substantial amount of input data, making it particularly useful in image processing. In the health industry, imaging is crucial in allowing health practitioners to visualize what is happening to a patient. As a result, DL has increased importance in health industries, assisting the process of analyzing medical images \cite{b7}. Recently, a Deep Learning image processing model developed to detect Alzheimer's disease is reported to be accurate 99.7\% of the time. \cite{b8}. 

\subsection{Bayesian Network}

\begin{figure}[htbp]
\centerline{\includegraphics{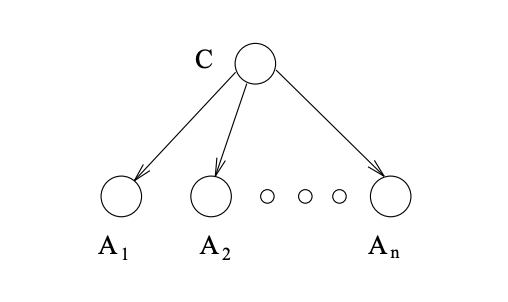}}
\caption{Sample Bayesian Network.}
\label{bn}
\end{figure}

Bayesian Network (BN) utilizes concepts from Bayes’ theorem, it updates the current hypothesis in the model based on newly observed information. \cite{b9}. BN is a graph that consists of nodes for each feature. Dependent nodes are connected via directed paths to indicate the relationship. This setup allows BN to predict data with uncertainty using predefined conditional probabilities. \cite{b10}. BN focuses on predicting data with uncertainty on conditional probabilities under prior beliefs. To construct a Bayesian Network, features can be classified as nodes with dependency links. Dependency links are arrows directed from one node to another, indicating the first node is dependent on the second. In Fig.~\ref{bn}, ``the classifier learns from training data the conditional probability of each attribute $A_i$ given the class label $C$, classification is then done by applying Bayes rule to compute the probability of C given the particular instance of $A_1, \cdots, A_n$, and then predicting the class with the highest posterior probability" \cite{b11}. The advantage of BN's ability to make accurate predictions with limited data over traditional machine learning algorithms has been adopted by many health institutes for research in cardiac conditions, cancer, psychological disorders, and lung disorders \cite{b12}.

\section{Methodologies}
Though both DL and BN models are applicable in medical industries, they each have their limitations and drawbacks. DL has solid abilities to classify a substantial amount of data including image classification, but does poorly in situations with limited data and high uncertainty. On the other hand, BNs are great at predicting under uncertainty but not necessarily efficient when handling large datasets, which may lead to creating an excessively massive graph. As a result, this article aims to analyze the possibility of combining the two models into one, aiming to maximize the profits while minimizing the drawbacks. 

\subsection{SWA-Gaussian Approach}\label{AA}
\begin{figure}[htbp]
\centerline{\includegraphics{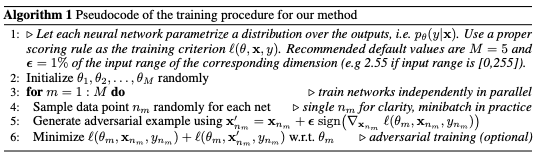}}
\caption{Algorithmic implementation of Deep Ensemble.}
\label{swag_alg}
\end{figure}

Reference \cite{b13} introduces Fig.~\ref{swag_alg}, the SWA-Gaussian (SWAG) approach to implement Bayesian characteristics into DL models. The Stochastic Weight Averaging procedure (SWA) is to obtain a solution $\theta_{SWA}$ to be the average of the weights $\theta_i$ after i epochs. And by denoting the diagonal to be $\Sigma_{diag} =$ diag$(\bar{\theta^2}-\theta^2_{SWA})$, we have set up a Gaussian distribution for $\tilde{\theta_i}$ with mean and standard deviation proportional to $\theta_{SWA}$ and $\Sigma_{diag}$ respectively, which can be used for a Bayesian Network. This transition, denoted SWAG, can allow a smooth combination of Deep Learning Networks and Bayesian Network to be made, enabling the ability to capture uncertainty with minimal modifications to the SWA training procedure. This approach can be adopted inside Deep Learning Models without significant modification to the code, to produce a Bayesian Deep Learning Model to predict more accurately under uncertainty.

\begin{figure}[htbp]
\centerline{\includegraphics{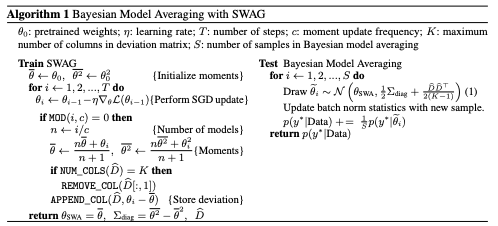}}
\caption{Algorithmic implementation of SWAG.}
\label{ensemble_alg}
\end{figure}

\subsection{Deep Ensemble}
Reference \cite{b14} introduces a method to combine Bayesian Network and Neural Network(NN) models by training multiple NN models by randomizing different initializations, as shown in Fig.~\ref{ensemble_alg}. All the trained models will be combined into an ensemble and viewed as an equivalent Bayesian Model. Although this is a simple approach to combine Bayesian and Deep Learning characteristics, the space and time complexity of this approach seems very chaotic. The construction and training of one Deep Learning Network is complicated enough to be extended into multiple networks to be trained as information for a Bayesian Model. However, with enough resources, this could also be a viable approach. 

\subsection{Bayesian Neural Network}
Reference \cite{b15} introduces Bayesian Neural Network (BNN) uses a concept of Variational Inference (VI). VI approximates the posterior with a distribution $q_\phi(\theta)$ by optimising a Monte Carlo estimate of the Evidence Lower Bound (ELBO):
\begin{equation}
\mathcal{L}(\phi) \approx \frac{1}{M} \sum^M_{m=1} \sum^N_{n=1}logp(y_n | \theta_m,x_n) - KL(q_\phi(\theta) ||p(\theta)) \label{eq}
\end{equation}
Then make predictions by sampling from $q_\phi(\theta)$:
\begin{equation}
p(y^*|x^*,\mathcal{D}) \approx \frac{1}{M}\sum^M_{m=1}p(y^*|x^*,\theta_m) \label{eq2}
\end{equation}
``A common choice for $q_\phi(\theta)$," described in reference \cite{b15}, is ``the mean-field Gaussian Approximation (MFVI)." 

\section{Analysis}
\begin{figure}[htbp]
\centerline{\includegraphics[width=0.5\textwidth]{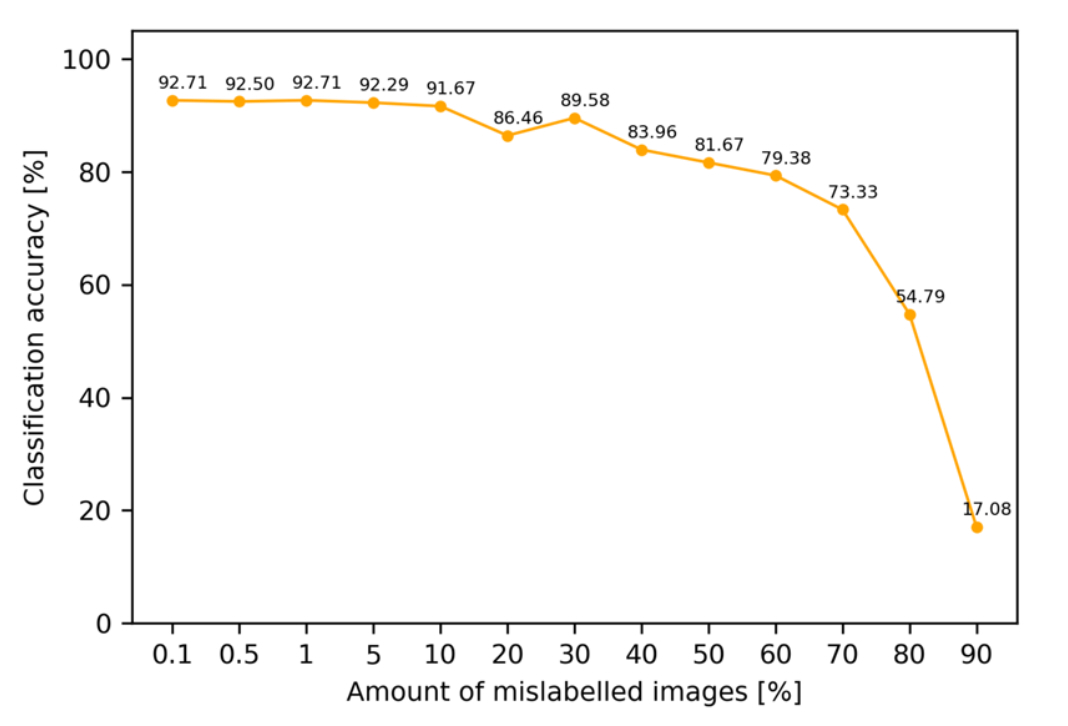}}
\caption{Accuracy of ARA-CNN with mislabeled images.}
\label{ann}
\end{figure}

There exist many approaches and methods to combine Bayesian Models and Deep Learning Models into one so-called Bayesian Deep Learning (BDL) Model or Deep Bayesian Learning Model. The combination of the two produces a more powerful predictive model that can not only account for large-scale datasets, but also handle data with uncertainty. Utilizing this improved model, the classification of complex images with uncertainty can now be achieved. In the modern day, imaging plays a crucial role in the diagnostic process of cancer development. Patients first arrange an initial assessment with General Practitioners (GPs) to present symptoms, GPs then refer the patient for a specialist consultation where the specialist will order imaging tests, such as X-rays, MRI, and CT-Scans, after the imaging results are available, the specialists then use the results to determine the severity of cancer and prescribe treatment accordingly \cite{b16}. Based on the process above, imaging interpretation skills are the key for specialists to unlock reliable diagnosis results and assign treatment, any inaccurate interpretation will result in mistreatment, which is devastating to patients who are already suffering from cancer. To maximize accuracy for the image interpretation process, the BDL model is a viable approach. Research using BDL models has shown excellent results so far, ``with over 98\% accuracy, one prototype and 2 data sets are used for clinical diagnosis and predictions for cancer and diabetes" \cite{b17}. An Accurate, Reliable, and Active Bayesian Convolutional Neural Network (ARA-CNN) is used to classify histopathological images of colorectal cancer \cite{b18}. Shown in Fig.~\ref{ann} from \cite{b19}, ARA-CNN maintains a high accuracy rate until the percentage of mislabeled images is relatively high. Similarly, roughly 2400 cheek mucosa images were collected as data for oral cancer research \cite{b20}. These images were labelled as "normal" and "suspicious" and given to a trained Bayesian Deep Neural Network for classification. The results show a good performance where the model has achieved 85.6\% accuracy. The research has also tested the model under low image quality (data with uncertainty) and the results indicated the model can produce informative uncertainty estimation \cite{b21}. Without the use of BDL in cancer diagnostics, an 11.1\% error rate was reported as the average diagnosis error rate for all cancer types \cite{b22}. Based on the observations above, BDL is considerably reliable to be used in the cancer imaging diagnosis process. 

\section{Conclusion}
In this paper, Deep Learning Models, Bayesian Models, and their benefits and limitations have been explored. Based on their unique characteristics, a Bayesian Deep Learning Model can be developed to obtain the advantages of both models while avoiding the disadvantages. There are multiple ways to combine both models effectively into one, however, the difference in the accuracy of the combination models is yet to be explored. Regardless of the implementation, a successfully designed Bayesian Deep Learning Model has shown success in the medical field with regard to image classification for cancers. This paper concludes that a well-designed Bayesian Deep Learning can be an effective approach to cancer diagnosis by accurately interpreting images for health practitioners. However, errors still exist in classification using the Bayesian Deep Learning approach and there will always be areas for improvement. Such improvements may include finding the best way to combine the models, selecting the best Deep Learning and Bayesian Model before the combination, or even combining a third type of machine learning model with the Bayesian Deep Learning Model, but these probable improvements methods will not be further investigated in this paper.

\end{document}